\documentclass[11pt]{article}

\usepackage[preprint]{acl}

\usepackage{times}
\usepackage{latexsym}
\usepackage{enumitem}
\usepackage{multirow}
\usepackage{multicol}
\usepackage{amsmath}
\usepackage{xcolor}
\usepackage{graphicx}
\usepackage{array}
\usepackage{booktabs}
\usepackage{subcaption}
\usepackage{soul}
\usepackage[most]{tcolorbox}
\usepackage{listings}
\usepackage{url}
\usepackage{amssymb}
\usepackage{hyperref}

\usepackage[T1]{fontenc}

\usepackage[utf8]{inputenc}

\usepackage{microtype}

\usepackage{inconsolata}

\usepackage{graphicx}

\title{MyAG: A Graph-Based Framework for Designing and Analyzing Composable LLM Agent Systems}

\author{
Zhisong Zhang\\
Department of Computer Science, City University of Hong Kong\\
\texttt{zhisong.zhang@cityu.edu.hk}
}

\begin{document}
\maketitle
\begin{abstract}

We present MyAG, a graph-based framework for designing and analyzing composable LLM agent systems. Our framework separates agent system construction into three graph abstractions: a component graph for agents, environments, and modules; a workflow graph for execution control; and a search graph for runtime execution. This separation allows users to flexibly reuse the same components with different strategies. We further support hierarchical composition through recursive system nodes and provide monitoring and visualization tools for inspecting agent execution. Experiments on representative agent applications show that our framework supports flexible agent system design and helps analyze performance-efficiency tradeoffs. Our framework is publicly available and fully open-source.\footnote{Our code is publicly available at \url{https://github.com/zzsfornlp/MyAG}, and a demo video is provided at \url{https://github.com/zzsfornlp/MyAG/blob/main/demo.mp4}.}

\end{abstract}

\section{Introduction}

With the development of large language models (LLMs), agent systems built upon them have seen rapid adoption in a wide range of applications \citep{wang2024survey,xi2025rise,luo2025large}. Building such systems usually involves various design aspects, including how different modules are composed, how the system workflow is executed, and how runtime search strategies are applied. These aspects are often closely related but serve for different purposes. It is useful to examine and modify them separately, since different applications and experiments may require different agent components, workflows, and search strategies \citep{li-2025-review,zhang2025aflow,shang2025agentsquare}.

\begin{figure}[t]
	\centering
	\includegraphics[width=0.47\textwidth]{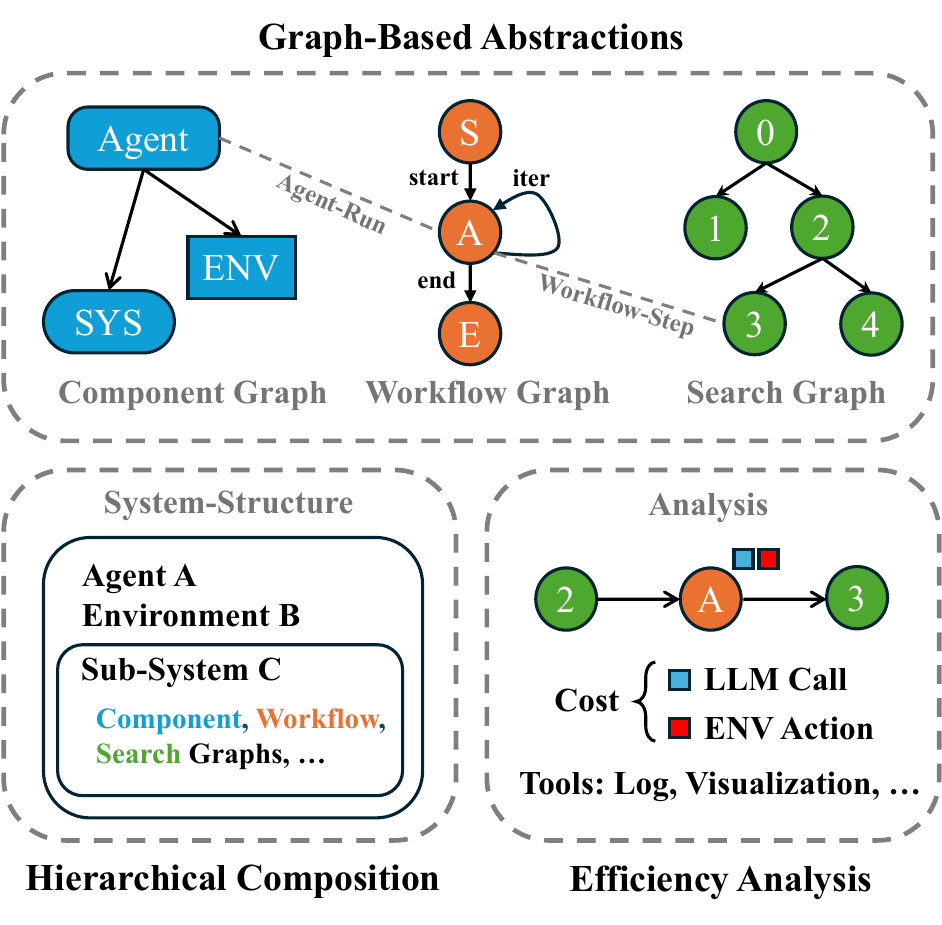}
	\caption{Overview of the core features of MyAG.}
    \vspace*{-5mm}
	\label{fig:overview}
\end{figure}

\begin{table*}[t]
\centering
\small
\begin{tabular}{l l ccccc}
\toprule
Framework 
& \#Core-Lines
& Workflow Support
& Search Support
& Hierarchical Systems 
& Efficiency Analysis \\
\midrule
\midrule
\href{https://github.com/langchain-ai/langgraph}{LangGraph} & ~~~--~~(\textasciitilde 27K) & \checkmark & -- & \checkmark & $\circ$\\
\href{https://github.com/microsoft/agent-framework}{Microsoft-Agent} & ~~~--~~(\textasciitilde 51K) & \checkmark & -- & \checkmark & $\circ$\\
\href{https://github.com/camel-ai/camel}{CAMEL} & ~~~$\circ$~~(\textasciitilde 18K) & $\circ$ & -- & $\circ$ & $\circ$\\
\href{https://github.com/foundationagents/metagpt}{MetaGPT} & ~~~$\circ$~~(\textasciitilde 25K) & $\circ$ & -- & $\circ$ & $\circ$\\
\href{https://github.com/microsoft/autogen}{AutoGen} & ~~~$\circ$~~(\textasciitilde 9K) & \checkmark & -- & $\circ$ & $\circ$\\
\href{https://github.com/huggingface/smolagents}{smolagents} & ~~~\checkmark~(\textasciitilde 6K) & -- & -- & \checkmark & $\circ$ \\
\href{https://github.com/wanxingai/LightAgent}{LightAgent} & ~~~\checkmark~(\textasciitilde 4K) & $\circ$ & $\circ$ & $\circ$ & $\circ$ \\
\midrule
\href{https://github.com/zzsfornlp/MyAG}{MyAG} (ours) & ~~~\checkmark~(\textasciitilde 4K) & \checkmark & \checkmark & \checkmark & \checkmark \\
\bottomrule
\end{tabular}
\caption{Comparison between MyAG and other representative open-source agent frameworks. ``\#Core-Lines'' provides an approximate number of lines in the core implementation and is used as a rough proxy for the size of the framework. For other feature entries, ``\checkmark'' indicates first-class support, ``$\circ$'' indicates partial support or support through user customization, and ``--'' indicates that the feature is not a primary focus.}
\label{tab:compare}
\end{table*}

Several existing frameworks and libraries have been proposed to support the development of LLM-based agent systems. For example, frameworks such as CAMEL~\citep{li2023camel}, MetaGPT~\citep{hong2024metagpt}, and AutoGen~\citep{wu2024autogen} provide abstractions for multi-agent collaboration; libraries such as LangGraph~\citep{langgraph2026github} and Microsoft Agent Framework~\citep{microsoft2026agentframework} support graph-based workflow orchestration; and runtime search strategies have been explored in research works such as tree-of-thought~\citep{yao2023tree} and tree-search-based agents~\citep{koh2025tree}. These systems have provided useful tools for building agent applications. However, many of them focus on specific aspects of agent system construction, and the boundaries between component composition, workflow control, and runtime search are often not made explicit. This makes it less convenient to conduct holistic agent evaluation under different workflows and search strategies, or analyze the efficiency of different agent designs~\citep{kapoor2026holistic}. In addition, some feature-rich frameworks introduce relatively large codebases and complex abstractions, while lightweight libraries such as smolagents~\citep{smolagents} and LightAgent~\citep{cai2025lightagent} mainly focus on simplifying agent construction and do not provide full support for complex workflow control, runtime search, and efficiency analysis.

In this work, we present MyAG (\textbf{My AG}ent), a lightweight software framework for building and analyzing LLM-based agent systems. MyAG introduces graph-based abstractions that separate component composition, workflow control, and runtime search. Specifically, the component graph describes agents, environments, and system modules; the workflow graph specifies how these components are executed; and the search graph records the runtime execution and search process. This design allows users to keep the same set of components while being able to switch among different workflows or search strategies in a flexible way.

Building on these graph abstractions, we further support flexible composition of agent systems. We introduce a special type of node, called a system node, as the main container of the component, workflow, and search graphs. Since a system can itself be used as a component inside another system, we support hierarchical construction of agent systems and encourages modular reuse. In addition, our framework supports agent efficiency analysis by recording different types of execution costs, including LLM-side costs, such as model calls and token usage, as well as environment-side costs, such as environment actions and action latencies. These measurements allow users to evaluate different workflows and search strategies in terms of both task performance and execution efficiency~\citep{yehudai2025survey}.

The main highlights of our proposed framework are summarized as follows:
\begin{itemize}
    \item We provide graph-based abstractions by separating component composition, workflow control, and runtime search. (\S\ref{sec:graphs})
    \item We support composable and hierarchical construction of agent systems. (\S\ref{sec:hier})
    \item We provide efficiency-aware analysis tools, enabling comprehensive comparison of different workflows and search strategies. (\S\ref{sec:eff})
    \item We provide monitoring and visualization tools for inspecting agent execution. (\S\ref{app:vis})
\end{itemize}

\section{Framework}

Figure~\ref{fig:overview} provides an overview of our framework. We focus on three design aspects: graph-based abstractions for component specification, workflow control, and runtime search; hierarchical composition of agent systems; and efficiency analysis based on execution records. Table~\ref{tab:compare} compares our framework with representative open-source LLM agent frameworks in terms of implementation size and supported features. In the following sub-sections, we will describe the details of these designs.

\subsection{Graph-Based Abstractions}
\label{sec:graphs}

We use three graphs to represent different structural aspects of an agent system: the \emph{component graph}, the \emph{workflow graph}, and the \emph{search graph}. The component graph specifies the available components in the system, such as agents, environments, and tools. The workflow graph indicates how these components are executed at runtime, including control flow and stopping conditions. The search graph records the detailed runtime execution process, such as states, branches, and trajectories. Analogous to a computer program, these three graphs can be viewed as describing variable declarations, program logic, and runtime states, respectively. In this sub-section, we use a web agent system as a typical running example to illustrate how these graphs are specified and used.

\begin{figure}[t]
	\centering
	\includegraphics[width=0.425\textwidth]{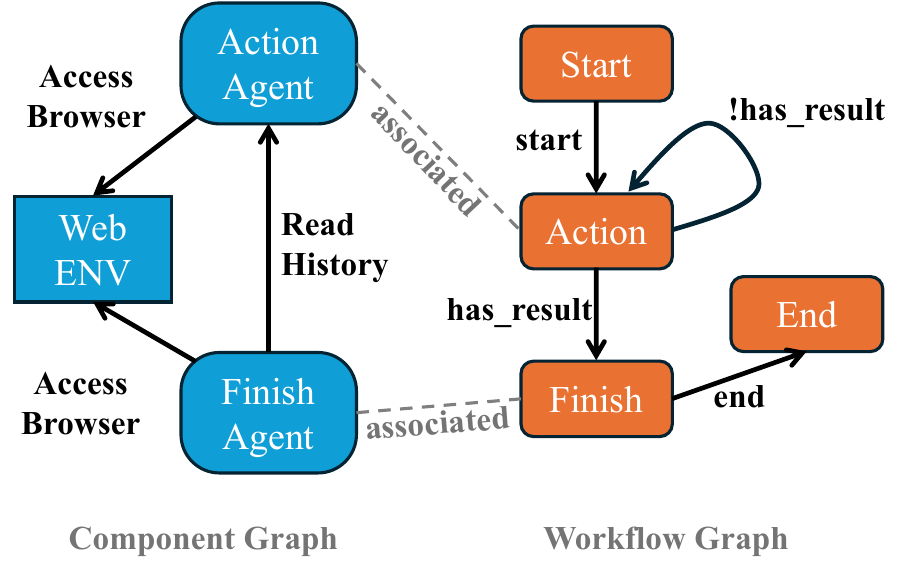}
    \vspace*{-3mm}
	\caption{Illustration of the component graph and the workflow graph of the web agent system.}
    \vspace*{-3mm}
	\label{fig:graphs}
\end{figure}

\paragraph{Component Graph.}
The component graph specifies what components an agent system contains. It mainly includes agent nodes, which represent LLM-based decision-making modules, and environment nodes, which represent non-LLM environments or tools that the agents interact with. It can also contain system nodes, which wrap another agent system as a reusable component; we discuss this hierarchical composition in \S\ref{sec:hier}. An edge in the component graph indicates that one component can access or interact with another component. For example, an agent node may call functions exposed by an environment node, or one agent node can read the memory or output of another agent node. Figure~\ref{fig:graphs} (left) shows the component graph of a typical web agent system. In this example, both the ``Action'' node and the ``Finish'' node can access the web browser environment, while ``Finish'' can also read the execution history of ``Action'' to produce the final formatted answer.

\paragraph{Workflow Graph.}
The workflow graph specifies how an agent system is executed to complete a target task. A common agent workflow follows the ReAct-style pattern~\citep{yao2023react}, which alternates between action decision and action execution. However, more flexible agent systems, especially multi-agent systems, often require more complex execution structures, such as branching, loops, and different stopping conditions. We represent such execution logic with a workflow graph.
Each workflow graph contains a start node and an end node. During execution, we maintains a workflow-level variable map that stores intermediate information shared across workflow nodes. A workflow node is usually associated with an agent node in the component graph. When such a workflow node is reached, the corresponding agent node is executed with its own input preparation and output parsing for the LLM call. After the execution, we determine the next workflow node according to the conditions specified on the current node's outgoing edges. Figure~\ref{fig:graphs} (right) shows the workflow graph of the web agent. After the ``Action'' node is executed, its provides an output variable ``has\_result'', which decides whether the next step returns to action decision or proceeds to the ``Finish'' node.

\paragraph{Search Graph.}
The search graph records the runtime execution process and supports the exploration of alternative execution paths. Many agent systems follow a greedy strategy, where the system proceeds along a single trajectory. However, prior work has shown that exploring alternative paths can be beneficial for complex tasks~\citep{yao2023tree,koh2025tree,li2025encompass}. We support different search strategies by storing runtime states as nodes in the search graph and allowing execution to resume from a selected node through state restoration. This design makes it possible to implement different search strategies such as greedy search, beam search, or best-first search under the same component and workflow definitions.
Some search strategies require additional support from the workflow and component graphs. For example, best-first search requires a scoring function for selecting the next node to expand. This scorer can be typically implemented as a specialized scoring node in the workflow graph, together with a corresponding scoring agent in the component graph. Typically, the scoring agent can use LLM-as-a-judge methods~\citep{zheng2023judging} to evaluate candidate nodes and provide scores to guide the search.

\vspace*{2mm}
Overall, the three-graph abstractions allow the decoupling of what components are available, how they are executed, and how runtime trajectories are explored. This separation makes it easier to reuse the same agent components under different workflows and search strategies, and provides a basis for systematic comparison and evaluation.

\subsection{Hierarchical Composition}
\label{sec:hier}

Many agent tasks can be naturally decomposed into sub-tasks that require different tools, environments, or execution strategies. For example, consider a long-horizon task that asks an agent to find a paper on the Internet and then analyze its content. One sub-task focuses on locating and downloading the paper through a web environment, while another sub-task focuses on reading and analyzing the downloaded file. Although a single agent system may handle both sub-tasks, a more modular design is to build a web sub-system and a file sub-system separately, and compose them in a higher-level system. These sub-systems may share the same underlying LLM, but they can use different tools, environments, and workflows.

\begin{figure}[t]
	\centering
	\includegraphics[width=0.45\textwidth]{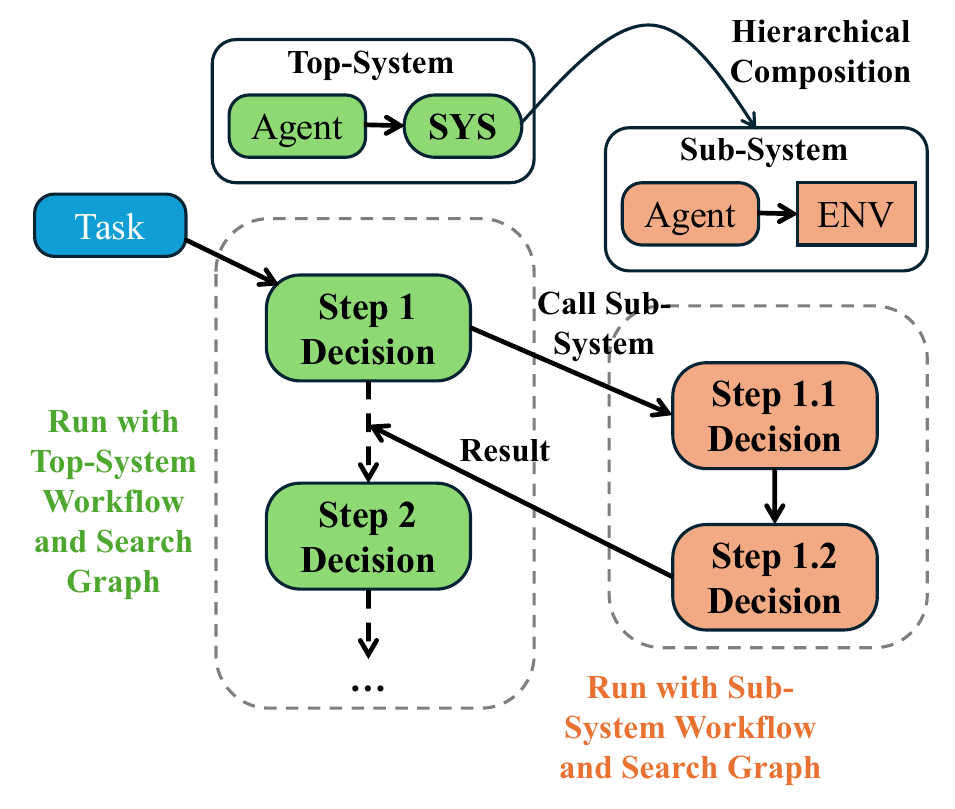}
    \vspace*{-3mm}
	\caption{Illustration of the hierarchical composition of a top-level system and its sub-system.}
    \vspace*{-3mm}
	\label{fig:hier}
\end{figure}

In our framework, we support such hierarchical composition through \emph{system nodes}. A system node is a special node in the component graph that wraps another complete agent system. The node ``\textbf{SYS}'' in Figure~\ref{fig:hier} shows an example of a system node. Each system, including both the top-level system and any sub-system, contains its own component graph, workflow graph, and search graph. Therefore, a sub-system is not only a tool wrapper, but can also maintain its own internal components and execution logic. This allows users to package a group of components and workflows as a reusable module, while still retaining fine-grained control over how each system is executed.

To allow a top-level system to use its sub-system, we expose the sub-system through function-call interfaces. We provide two modes for managing the state of a sub-system across calls. In the stateless mode, the sub-system is reset before each call and behaves similarly to a tool. This is useful when the sub-system is expected to solve an independent sub-task given the current input. In the stateful mode, the sub-system keeps its own memory and runtime state across calls, allowing it to cooperate with the top-level system in a more persistent multi-agent manner. These two modes provide a simple way to reuse sub-systems under different scenarios.

\subsection{Efficiency Analysis}
\label{sec:eff}

Efficiency is an important consideration for LLM-based agent systems, especially when agents require many model calls or interact with slow external environments. However, compared with task performance, execution cost is often less systematically analyzed, and there is not yet a single widely adopted protocol for measuring agent efficiency~\citep{yehudai2025survey,kapoor2025ai}. In practice, agent efficiency can depend on multiple factors, including the number of LLM calls, token usage of each call, and action latencies. We therefore provide execution-level tools to record and analyze these costs.

In our framework, we mainly consider two types of costs: the LLM inference cost ($C_L$), and the environmental interaction cost ($C_E$). Given an agent trajectory $\tau = (s_0, a_0, s_1, a_1, \dots, s_T)$, these costs can be computed from the execution records stored during runtime.

The LLM inference cost $C_L$ measures the cost of the agent's internal decision process. We define it as the weighted sum of input and output tokens used by LLM calls along the trajectory: $C_L = \sum_{t=0}^{T-1} (\alpha \cdot L_{t}^{\mathrm{in}} + \beta \cdot L_{t}^{\mathrm{out}})$, where $L_t^{\mathrm{in}}$ and $L_t^{\mathrm{out}}$ denote the number of input and output tokens at step $t$, respectively. The coefficients $\alpha$ and $\beta$ are weighting factors for input and output tokens. They can be set according to API prices, empirical inference latency, or other user-defined cost models. This makes the metric adaptable to different model providers and deployment settings.

The environmental interaction cost $C_E$ measures the cost of executing actions in external environments. This cost is important because environment actions may have different latencies or resource requirements. For example, in a web environment, scrolling a page is usually cheaper than loading a new page. We define it as: $C_E = \sum_{t=0}^{T-1} w(s_t, a_t)$, where $w(s_t, a_t)$ is a user-defined or empirically measured cost function for taking action $a_t$ at state $s_t$. In our implementation, this cost is instantiated by empirically measured action latency.

These two costs enable the analysis of agent executions from both the model side and the environment side. Together with task performance, these measurements support performance-efficiency trade-off analysis across different workflows and search strategies.


\section{Evaluation}

\begin{figure*}[t]
	\centering
	\begin{subfigure}[b]{1.0\textwidth}
		\includegraphics[width=0.975\textwidth]{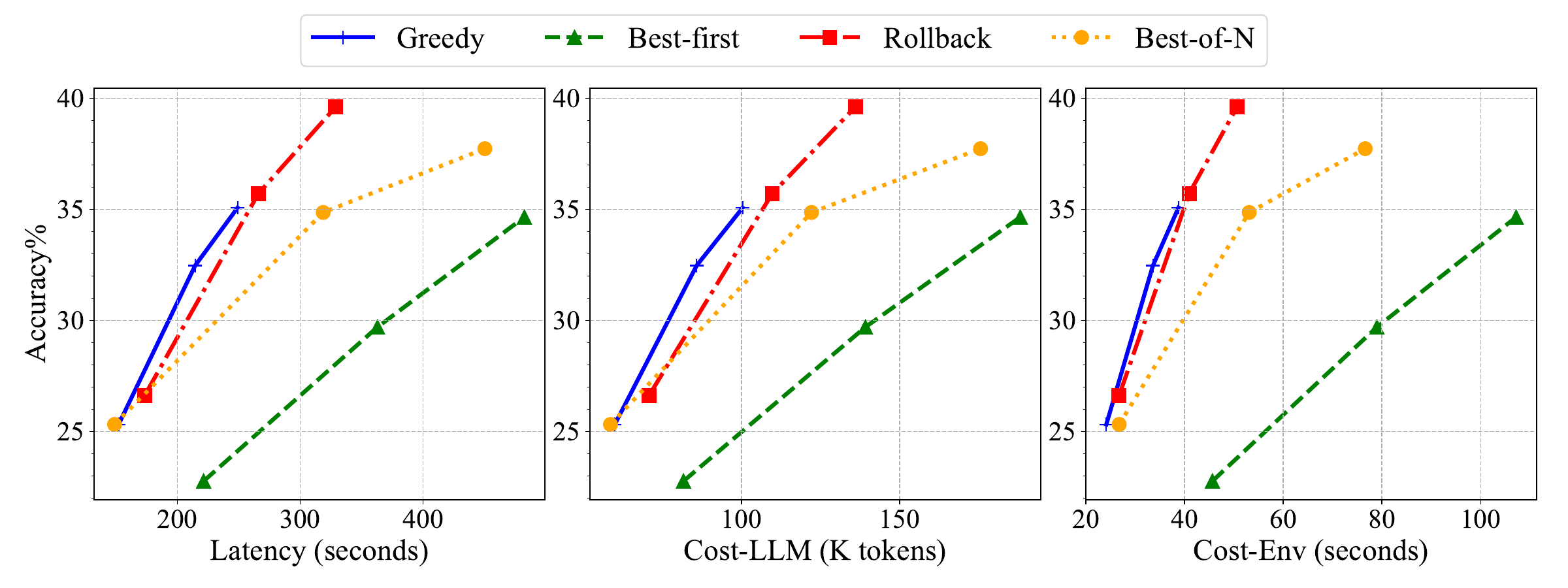}
	\end{subfigure}
	\vspace*{-5mm}
	\caption{Performance-efficiency analysis of different strategies (on GAIA-text). For each strategy, we report task performance and three instance-level efficiency metrics: overall latency in seconds, LLM inference cost measured by the number of tokens in thousands, and environmental interaction cost measured by cumulative interaction latency. All efficiency metrics are averaged over evaluation instances.}
	\label{fig:res1}
\end{figure*}

\subsection{Settings}

\paragraph{Setup} We evaluate our framework on two representative agent tasks: complex question answer, using the text subset of the GAIA benchmark~\citep{mialon2024gaia}, and web navigation, using Mind2Web-Live~\citep{pan2024webcanvas}. In our main experiments, we use \textsc{Qwen3.5-27B} as the primary underlying LLM and implement the automatic web browser environment with Playwright.\footnote{\url{https://playwright.dev/python/}} Each experiment is run three times and the averaged performance is reported, which could provide more stable results.


\paragraph{Comparisons} We compare several workflow and search strategies supported by our framework: 1) The \emph{Greedy} strategy follows a ReAct-style execution pattern~\citep{yao2023react}, where the agent perform one action at each step based on the latest state; 2) The \emph{Best-first} strategy maintains a search tree and expands the highest-scored node at each step~\citep{koh2025tree}; 3) The \emph{Rollback} strategy allows the agent to decide whether to return to a previous state before taking the next action~\citep{zhang-etal-2026-webrollback}; and 4) The \emph{Best-of-N} strategy performs multiple parallel runs and uses an additional LLM judge to select the best trajectory~\citep{lightman2024lets}. Unless otherwise specified, we set the maximum step budget to 36, and we also vary this budget in our evaluation to study how different strategies scale with test-time computation.

\begin{figure*}[t]
	\centering
	\begin{subfigure}[b]{1.0\textwidth}
		\includegraphics[width=0.975\textwidth]{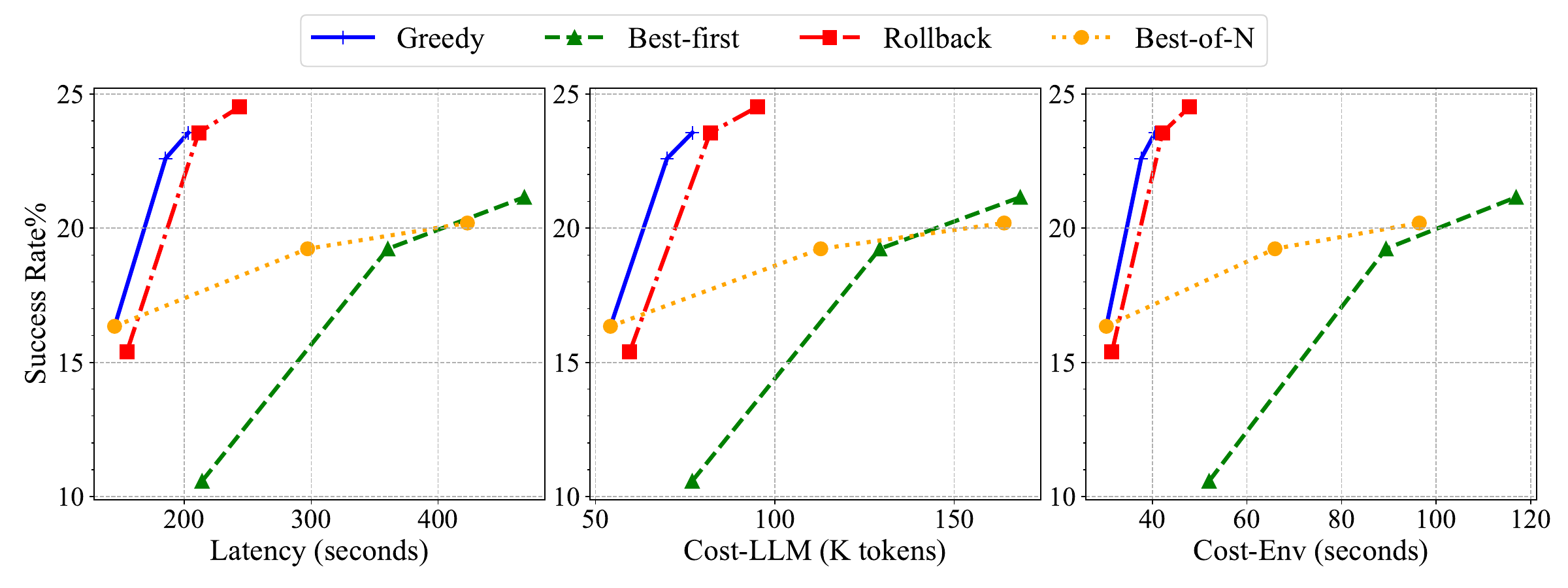}
	\end{subfigure}
	\vspace*{-5mm}
	\caption{Performance-efficiency analysis of different strategies (on Mind2Web-Live).}
	\label{fig:res2}
\end{figure*}

\subsection{Results}

The main results are shown in Figure~\ref{fig:res1} and \ref{fig:res2}, where different strategies show different patterns. Overall, the simplest \emph{Greedy} strategy is the most efficient in terms of computation usage, since it follows a single execution trajectory without additional scoring, rollback, or selection steps. The \emph{Rollback} strategy obtains the best overall performance when sufficient budget is available, but it is slightly less efficient than \emph{Greedy} due to the additional rollback decision and the more complex prompt. The \emph{Best-first} and \emph{Best-of-N} strategies can improve performance with more computation, but they also introduce larger overheads. To be noted, the effectiveness of some complex strategies depends on the quality of their verification modules, such as the final selector in \emph{Best-of-N} or the state-value scorer in \emph{Best-first}. In our experiments, we use the same policy LLM with specialized prompts for these modules, while stronger verification models or task-specific scorers may further improve these strategies. Our framework makes such comparisons straightforward, since different workflow and search strategies can be implemented and evaluated under the same framework.

\begin{table}[t]
\centering
\small
\begin{tabular}{c|c @{\hspace{5pt}} c @{\hspace{5pt}} c @{\hspace{5pt}} c @{\hspace{5pt}} c}
\toprule
Strategy & Action & Rollback & Finish & ENV & Reset\\
\midrule
Greedy & 78.6\% & - & \phantom{0}6.3\% & 13.8\% & \phantom{0}1.3\% \\
Best-first & 74.9\% & - & \phantom{0}2.9\% & 10.8\% & 11.4\% \\
Rollback & 75.6\% & 4.8\% & \phantom{0}4.8\% & 11.8\% & \phantom{0}3.0\% \\
Best-of-N & 72.2\% & - & 11.8\% & 13.8\% & \phantom{0}2.2\% \\
\bottomrule
\end{tabular}
\caption{Latency breakdown on different components, including LLM calls for the Action, Rollback, and Finish modules, environment interaction (ENV), and Reset for initialization or context switching.}
\label{tab:comp}
\end{table}

\begin{table}[t]
\centering
\small
\begin{tabular}{c | c  c  c}
\toprule
Action & Call\% & Time\% & Cost (s)\\
\midrule
click & 31.4\% & 43.5\% & 3.66\\
scroll & 29.3\% & 11.0\% & 1.00\\
type & 20.3\% & 19.5\% & 2.55\\
goto & 11.3\% & 13.5\% & 3.17\\
goback & \phantom{0}5.2\% & \phantom{0}4.7\% & 2.41\\
wait & \phantom{0}2.5\% & \phantom{0}7.8\% & 7.82\\
\bottomrule
\end{tabular}
\caption{Action-level cost analysis of the web agent system. ``Call\%'' denotes the percentage of action calls, ``Time\%'' denotes the percentage of action execution time, and ``Cost'' denotes the average latency per call.}
\label{tab:action}
\end{table}

\subsection{Analysis}

\paragraph{Component Cost}
We first break down the overall latency by component for different strategies, and the results are shown in Table~\ref{tab:comp}. \emph{Best-first} spends a much larger portion of time on environment resetting, reflecting its frequent context switching. Although \emph{Rollback} introduces an additional module for rollback decisions, this module is not triggered frequently and therefore does not introduce substantial overhead. \emph{Best-of-N} spends more time on the Finish module, since multiple parallel trajectories are executed and then compared before producing the final answer.

\paragraph{Action Cost} 
We further break down the latency cost of environment actions, and the results are shown in Table~\ref{tab:action}. The main motivation for recording action-level costs is that different environment actions can have substantially different latency profiles. This difference is clearly reflected in our analysis. For example, the ``scroll'' action is much cheaper than other actions: although it accounts for nearly 30\% of action calls, it contributes only around 11\% of the total environment interaction time. In contrast, the ``wait'' action is the most expensive action, which explains why its time percentage is much higher than its call percentage.

\paragraph{Underlying LLMs}
Finally, we evaluate different underlying LLMs to study how model capability affects agent performance and efficiency. We use models from the \textsc{Qwen3.5} family and vary the model size. Results are shown in Appendix~\ref{app:extra}.


\section{Conclusion}

We presented MyAG, a lightweight graph-based framework for designing and analyzing composable LLM agent systems. It separates agent system construction into component graphs, workflow graphs, and search graphs, making it easier to reuse components under different execution and search strategies. It also supports hierarchical composition through system nodes and provides monitoring and analysis tools for inspecting agent execution. Experiments and evaluations show that our framework can support flexible system design and help analyze performance-efficiency tradeoffs.

\section*{Limitations}

MyAG is designed as a lightweight, research-oriented framework rather than a full production platform. Therefore, it does not currently focus on production-level features such as distributed deployment, access control, or fault tolerance. In addition, the efficiency metrics used should be viewed as configurable analysis tools rather than standardized evaluation protocols, since the cost weights may depend on many factors, such as model providers and the hardware. Finally, our experiments aim to demonstrate the capability of the framework through representative agent applications. Broader evaluation across more environments and extensions to additional application scenarios are left to future work.

\newpage

\bibliography{main}

\clearpage

\appendix

\section{Monitoring and Visualization Tools}
\label{app:vis}

We provide monitoring and visualization tools for inspecting agent execution. The dashboard-style interface shows the step-by-step running process, including workflow transitions, agent calls, environment actions, intermediate outputs, and runtime metrics. This allows users to better understand agent behavior and analyze the trajectory of agent execution.

\begin{figure*}[t]
	\centering
	\includegraphics[width=0.975\textwidth]{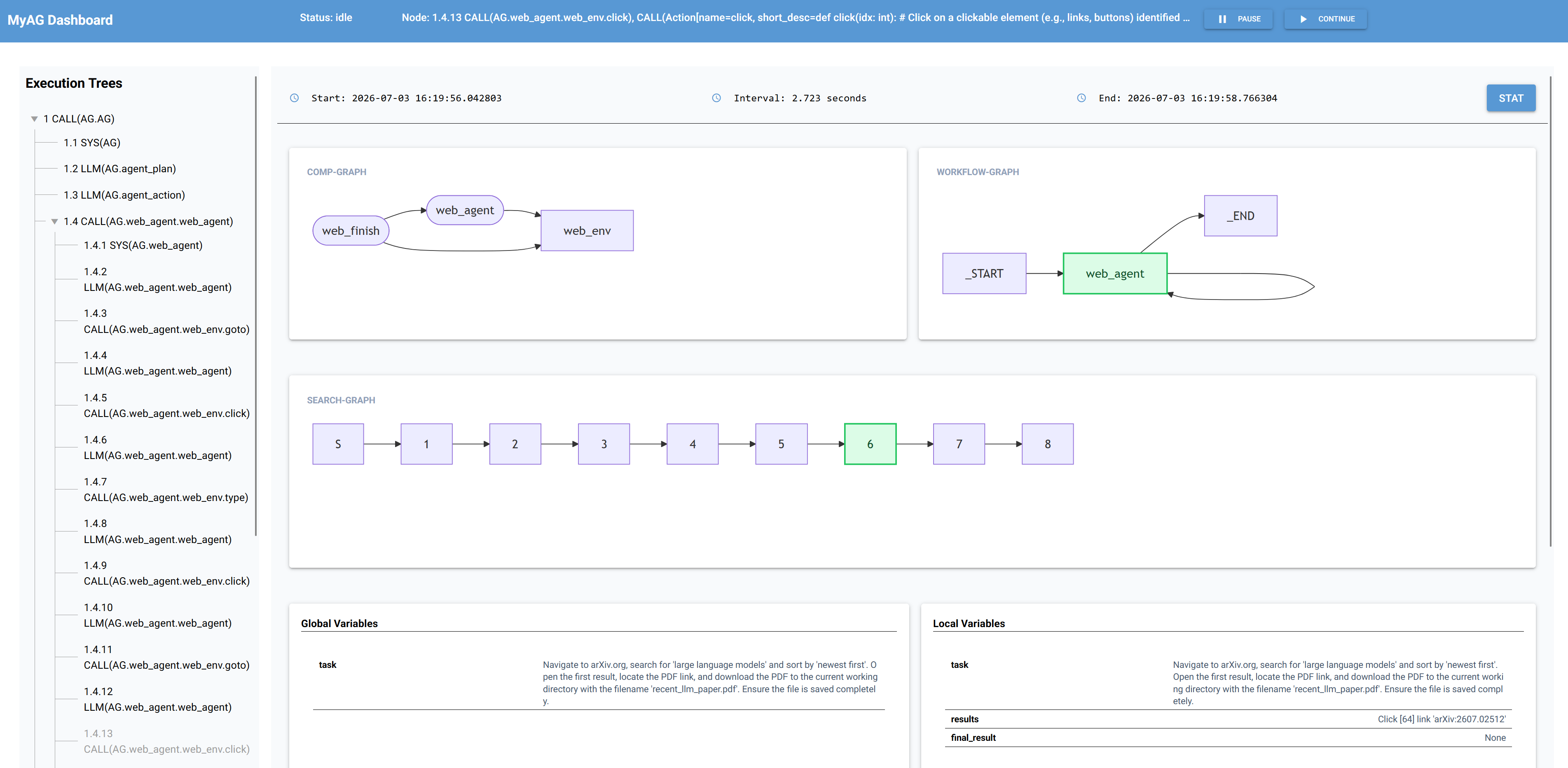}
    \vspace*{-3mm}
	\caption{A screenshot of the visualization dashboard.}
    \vspace*{-3mm}
	\label{fig:ui}
\end{figure*}

The dashboard interface is shown in Figure~\ref{fig:ui}. On the left column, we display the system execution trace, which is arranged in tree structures, where each node represents a key execution step, such as an LLM call or an environmental action call. This hierarchical tree structure also supports navigation across nested systems, making it easier to inspect the execution traces of hierarchically composed agent systems. On the right panel, we show the detailed information of the selected execution node, which can be chosen by clicking a node in the trace tree. The information includes the running time information, the visualization of the component graph, the workflow graph, and the search graph of the current agent system. In addition, the workflow-level global variables and node-level local variables are shown below the graph visualizations. The dashboard also supports cost visualization, including pie charts for latency distribution and token usage.

Overall, this toolkit provides a practical interface for monitoring and analyzing agent execution. It allows users to inspect both high-level execution traces and fine-grained runtime details, which is useful for debugging agent behavior and understanding the characteristics of different strategies.

\section{Extra Results}
\label{app:extra}

\begin{figure}[t]
	\centering
	\includegraphics[width=0.47\textwidth]{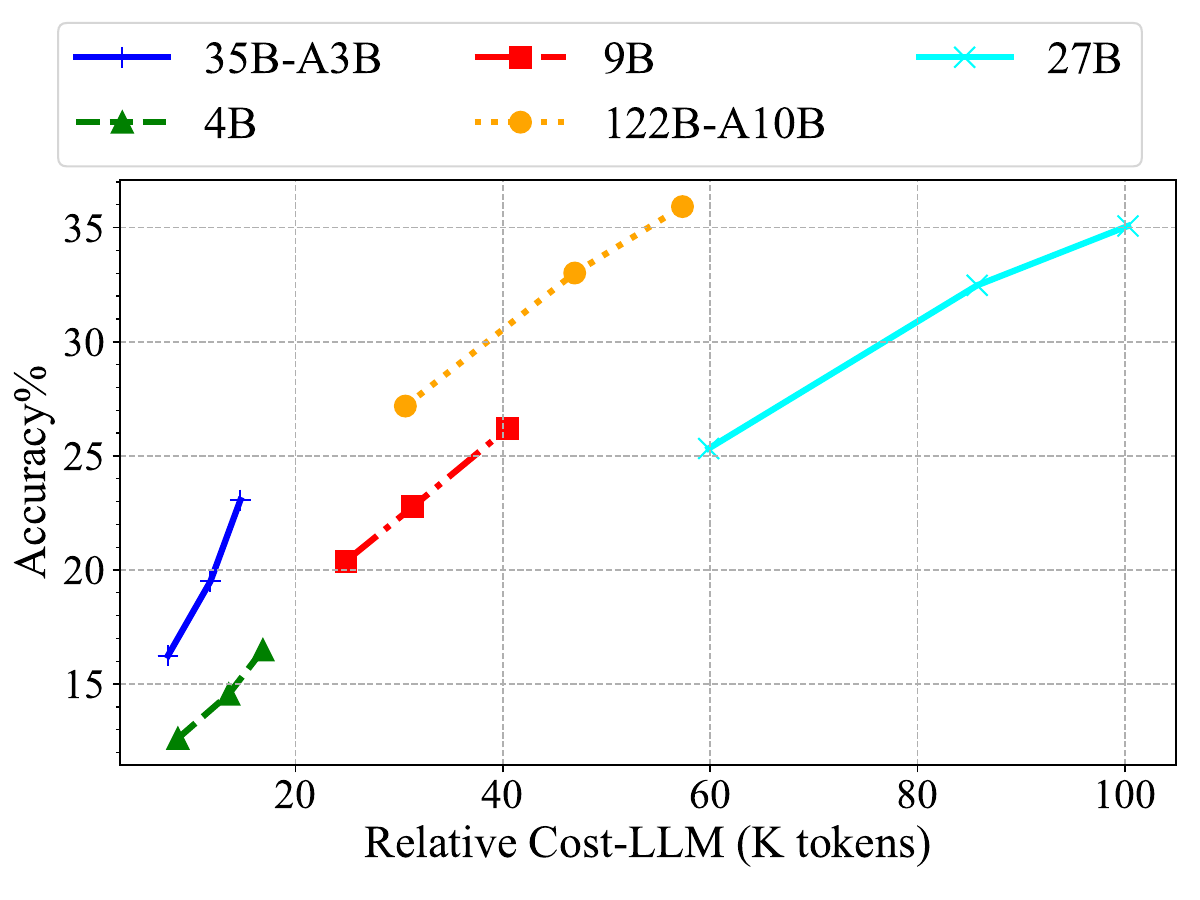}
	\caption{Performance vs relative token cost with different underlying LLMs.}
    \vspace*{-5mm}
	\label{fig:resB2}
\end{figure}

In the main experiments, we use \textsc{Qwen3.5-27B} as the underlying LLM and compare different workflow and search strategies. In this analysis, we instead vary the underlying LLM to investigate how model capability affects agent performance and efficiency. We consider models from the \textsc{Qwen3.5} family\footnote{\url{https://huggingface.co/collections/Qwen/qwen35}} with different sizes, including \textsc{Qwen3.5-4B}, \textsc{Qwen3.5-9B}, \textsc{Qwen3.5-27B}, \textsc{Qwen3.5-35B-A3B}, and \textsc{Qwen3.5-122B-A10B}. We evaluate these models on GAIA-text using the simple \emph{Greedy} strategy.

\begin{figure*}[t]
	\centering
	\begin{subfigure}[b]{1.0\textwidth}
		\includegraphics[width=0.975\textwidth]{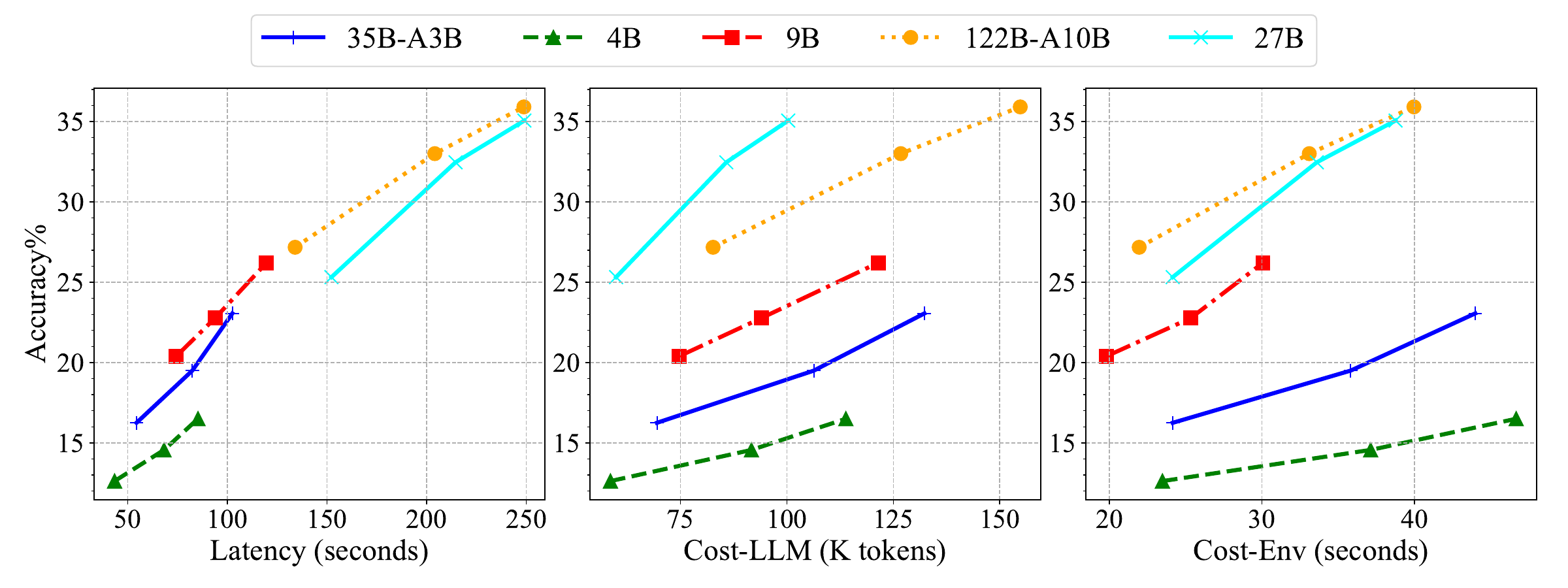}
	\end{subfigure}
	\vspace*{-5mm}
	\caption{Performance-efficiency analysis with different underlying LLMs.}
	\label{fig:resB1}
\end{figure*}

The results are shown in Figure~\ref{fig:resB1}. Smaller dense models have lower per-call computation but may require more interaction steps or fail more frequently, while larger models can provide stronger task performance at higher resource requirements. Directly comparing token usage across models does not fully reflect inference cost, since the cost of processing each token varies with model size and architecture. We therefore additionally consider a simplified cost model in which the per-token inference cost is assumed to be proportional to the number of activated parameters. Figure~\ref{fig:resB2} presents the results under this relative cost estimate.

Under this cost model, the mixture-of-experts (MoE) models provide better performance-efficiency tradeoffs, since only a subset of their parameters is activated for each token. Nevertheless, this estimate does not account for all deployment costs. In particular, such models may still require substantial GPU memory to store the full model parameters, which is not captured by the activated-parameter-based cost model.

Overall, these results suggest that the choice of the underlying LLM can substantially affect the performance-efficiency tradeoff of an agent system. There is no single model that is optimal under all deployment constraints: the best choice depends on the available inference budget, memory capacity, latency requirements, and target performance. This analysis also demonstrates how our framework can combine execution records with configurable cost models to compare agent systems across different underlying LLMs.

\end{document}